\pdfoutput=1
\documentclass[shortlabels]{sigchi-ext}
\usepackage[T1]{fontenc}
\usepackage{textcomp}
\usepackage[scaled=.92]{helvet} 
\usepackage{graphicx} 
\usepackage{balance}  
\usepackage{booktabs} 
\usepackage{ccicons}  
\usepackage{ragged2e} 

\usepackage[normalem]{ulem}
\useunder{\uline}{\ul}{}

\usepackage{enumitem}
\usepackage{multirow}

\def\plaintitle{LLMs for XAI: Future Directions for Explaining Explanations} \def\plainauthor{Alexandra Zytek, Sara Pido, Kalyan Veeramachaneni}

\def\plainkeywords{Explainable Artificial Intelligence (XAI), Large Language Models (LLMs), ML Explanations, Prompt Design, Interpretability, Human-Computer Interaction, Natural Language}

\title{LLMs for XAI: Future Directions for Explaining Explanations}

\numberofauthors{3}

\author{%
  \alignauthor{%
    \textbf{Alexandra Zytek\footnotemark[1],Sara Pidò\footnotemark[1], Kalyan Veeramachaneni}\\
    \affaddr{MIT}, \affaddr{Cambridge}, \affaddr{MA}, \affaddr{US}
    \email{zyteka@mit.edu, sarapid@mit.edu, kalyan@csail.mit.edu}
   }
   \footnotetext{$^1$These authors contributed equally to this work.}
}

\definecolor{linkColor}{RGB}{6,125,233}
\hypersetup{%
  pdftitle={\plaintitle},
  pdfauthor={\plainauthor},
  pdfkeywords={\plainkeywords},
  bookmarksnumbered,
  pdfstartview={FitH},
  colorlinks,
  citecolor=black,
  filecolor=black,
  linkcolor=black,
  urlcolor=linkColor,
  breaklinks=true,
}

\begin{document}


\copyrightinfo{Presented at The 2024 ACM CHI Workshop on Human-Centered Explainable AI (HCXAI)}

\maketitle
\RaggedRight{} 

\begin{abstract}
In response to the demand for Explainable Artificial Intelligence (XAI), we investigate the use of Large Language Models (LLMs) to transform ML explanations into natural, human-readable narratives. Rather than directly explaining ML models using LLMs, we focus on refining explanations computed using existing XAI algorithms. We outline several research directions, including defining evaluation metrics, prompt design, comparing LLM models, exploring further training methods, and integrating external data. Initial experiments and  user study suggest that LLMs offer a promising way to enhance the interpretability and usability of XAI.
\end{abstract}




\section{Introduction}
\label{sec:intro}

In recent years, the demand for Explainable Artificial Intelligence (XAI) has increased, driven by concerns about the opacity and lack of interpretability of complex machine learning models~\cite{arrieta2020explainable, bhatt_explainable_2020}. 
However, domain experts without ML experience still struggle to understand and use many ML explanations, which are often presented in a way that is not natural and human-readable \cite{bhatt_explainable_2020, jiang_two_2021, nyre-yu_considerations_2021, yang2023survey, zytek_sibyl_2021}. 

Having proven useful in a variety of domains~\cite{chang2023survey}, Large Language Models (LLMs) offer a promising way to advance the field of XAI. In this paper, we investigate whether LLMs can improve explanations by transforming them into natural, human-readable narratives. In particular, LLMs' access to extensive background information may allow them to further ``explain explanations'' by offering context information.

LLMs may aid in XAI in many ways \cite{wuUsableXAI102024}. For example, some work has investigated using LLMs to directly explain ML models \cite{bhattacharjeeLLMguidedCausalExplainability2024, kroeger2023large} or to understand user questions to generate the appropriate explanation \cite{nguyen2023black, shenConvXAIDeliveringHeterogeneous2023, slack2023explaining}. In this work, we instead focus on using LLMs to transform existing ML explanations, generated by theoretically-grounded explanation algorithms such as SHAP \cite{lundberg2017unified}, into natural language narratives. 
We begin by proposing research directions we believe will further this goal, and then discuss our current progress on a few of these directions.

\newcommand{\accuracy}{soundness}
\newcommand{\fluency}{fluency}
\newcommand{\completeness}{completeness}
\newcommand{\contextawareness}{context-awareness}
\newcommand{\Accuracy}{Soundness}
\newcommand{\Fluency}{Fluency}
\newcommand{\Completeness}{Completeness}
\newcommand{\Contextawareness}{Context-awareness}

\section{Research Directions}
Here we list several directions to investigate and improve LLMs for generating explanation narratives.

\textbf{R1. Defining evaluation metrics: } To allow researchers to compare narrative explanations from different models and techniques, formal metrics for evaluation must be defined.
Existing work \cite{xieInterpretationQualityScore2022a, zhouEvaluatingQualityMachine2021} proposes metrics for evaluating XAI explanations in general; in this paper, we propose an initial set of metrics adapted specifically for LLM-generated natural-language explanations. 

\textbf{R2. Prompt Design:} 
It is worth investigating how well LLMs are able to generate explanation narratives in a zero-shot manner. 
Prompt design involves crafting specific prompts to guide language models in producing nuanced and contextually varied explanations. This contributes to the robustness of the evaluation by mitigating the potential bias introduced by a single prompt. In this paper, we try five prompts on pretrained LLMs (particularly, GPT-3.5 and GPT-4).

\newcounter{PCounter}
\newcommand{\PLabel}[1]{%
    \refstepcounter{PCounter}
    \label{#1}
    P\thePCounter
}

\begin{table*}[t!]
\footnotesize
\caption{The prompts used for our initial experiments. Recurring components are at the top, followed by the five prompts. The context description in [INTRO] and the values in [EXP] varied by dataset and input. P\ref{prompt:fluent} and P\ref{prompt:context} were designed to address shortcomings in responses from P\ref{prompt:basic} (a lack of \fluency{} or \contextawareness{} respectively). P\ref{prompt:novals} and P\ref{prompt:vals} show examples of fulfilling specific user needs through prompt tuning.} 
\begin{tabular}{cp{0.9\textwidth}}
\toprule
\textbf{Code} &
  \textbf{Recurring Component} \\ \midrule
  {[}INTRO{]} &
   You are helping users understand an ML model's prediction. The model predicts house prices. \\ \addlinespace
{[}TASK{]} &
  I will give you feature contribution explanations, generated using SHAP, in (feature, feature\_value, contribution, average\_feature\_value) format. Convert the explanations in simple narratives. Do not use more tokens than necessary. \\ \addlinespace
 {[}EXP{]} & (Second floor square feet, 854, 12757.83, 583.0), (Construction date, 2003, 9115.72, 1976.1), (Basement area in sq. ft., 856, -6157.86, 1073.6) \\ \midrule
\textbf{Code} &
  \textbf{Prompt} \\ \midrule
\PLabel{prompt:basic} &
  {[}INTRO{]}{[}TASK{]}{[}EXP{]} \\ \addlinespace
\PLabel{prompt:fluent} &
  You are helping users who do not have experience working with ML understand an ML model's predictions. {[}TASK{]}. Make your answers sound as natural as possible, as though said in conversation. {[}EXP{]} \\ \addlinespace
\PLabel{prompt:context} &
  {[}INTRO{]}{[}TASK{]} Include context on how the specific instance compares to the average, and ensure the explanation highlights the significance of each feature's contribution to the overall prediction. {[}EXP{]}  \\ \addlinespace 
\PLabel{prompt:novals} &
  {[}INTRO{]}{[}TASK{]} Do not explicitly mention any feature or contribution values in your response. {[}EXP{]} \\ \addlinespace
\PLabel{prompt:vals} &
  {[}INTRO{]}{[}TASK{]} Be sure to explicitly mention all feature and contribution values in your response. {[}EXP{]} \\ \bottomrule  
\end{tabular}
\label{tab:prompts}
\end{table*}


\begin{table*}[ht]
    \centering
        \caption{Metrics used to rate explanations. As indicated by arrows, for \accuracy{}, \fluency{}, and \contextawareness{}, we assume higher values are better. For length, lower values are better. For \completeness{}, a score of 1 or 2 may be preferred depending on the target audience.}
    \resizebox{0.97\textwidth}{!}{%
    \begin{tabular}{|p{0.3\textwidth}|p{0.32\textwidth}|p{0.32\textwidth}|p{0.3\textwidth}|}
        \hline
        \textbf{Metric} & \textbf{0} & \textbf{1} & \textbf{2} \\
        \hline
        \textbf{\Accuracy{} \textuparrow :} The correctness of the information included in the narrative.  &
        The narrative includes one or more objective errors, such as reporting an incorrect contribution. & The narrative includes one or more misleading statements. For example, referring to a rating of 5 out of 10 as ``high''. & The narrative contains no errors. \\
        \hline
        \textbf{\Fluency{} \textuparrow :} The extent to which the narrative sounds ``natural'' or like it was generated by a human peer in conversation. & The narrative is very unnatural or confusingly worded, or directly repeats the given feature names and values without any narrative flow. & The narrative is somewhat natural. & The narrative sounds very natural, as though written by a human peer with good language skills. \\
        \hline
        \textbf{\Completeness{} \textendash :} The amount of information included in the narrative. & Some features given were missed entirely, or the direction of their contributions was not specified. & All features were described, but exact feature or contribution values were not given. & All feature values are given, and all contributions are described with directions and either exact values or descriptions (``contributed slightly''). \\
        \hline
        \textbf{\Contextawareness{} \textuparrow :} The degree to which the narrative ``explains the explanation'' by providing external context. & No context information is provided. & The answer includes references and comparisons to the average/mode feature values. & The answer includes further explanations of what may cause a specific contribution. \\
        \hline
        \textbf{Length} \textdownarrow & \multicolumn{3}{|p{0.9\textwidth}|}{The number of words in the response.} \\
        \hline
        
    \end{tabular}%
    }
    \label{tab:explanation_metrics}
\end{table*}

\textbf{R3. Comparing LLMs:}
Burton et. al. \cite{burton2023natural} demonstrated the effectiveness of smaller language models (T5 and BART) in generating explanation narratives. A thorough analysis comparing the effectiveness of newer, larger models in generating explanation narratives will contribute to a deeper understanding of their strengths and weaknesses.

\textbf{R4. Further training and finetuning:}
Explanation narratives generated by LLMs can be improved through the use of training methods such as model finetuning. For example, Burton et. al. \cite{burton2023natural} employed eight computer scientists to create a dataset of narrative versions of ML explanations. We propose furthering this work through a similar dataset generated by a larger group of non-technical experts, who may generate more broadly applicable samples.

\textbf{R5. Integration of Training and Other External Data:}
Information from the model's training data, as well as other external information about the domain such as guides or textbooks, can be integrated into the explanation generation process to get more context-aware explanations using techniques such as Retrieval-Augmented Generation (RAG). Guo et. al. \cite{guoAdaptingPromptFewshot2023} demonstrated the effectiveness of this approach for generating natural-language descriptions of tables; extending this concept to ML explanations may contribute to more comprehensive narratives.


We hope these directions provide a roadmap for further investigation into using LLMs for XAI. In the next section we introduce our current progress in some of these directions.

\section{LLMs for XAI: First Steps}

\label{sec:experimental_setup}
We began by investigating how well LLMs can transform explanations into narratives in a zero-shot manner, without any additional training, using GPT-3.5 and GPT-4. To begin with, we are focusing on SHAP explanations as inputs.

We applied a technique called prompt design, or crafting prompts to guide the language models in producing accurate and useful explanations. At this time, we have experimented with five prompts, shown in Table \ref{tab:prompts} (\textbf{R2}). 

To empirically compare the results from different prompts and models (and in the future, additional training methods), we adapted the metrics summarized by Zhou et. al. \cite{zhouEvaluatingQualityMachine2021} to evaluate XAI explanations (\textbf{R1}). As proposed in this work, we use the metrics of \textit{\completeness{}} and \textit{\accuracy{}} to measure the degree to which the narrative accurately represents the original explanation. We replace the term clarity with the more natural-language-specific term \textit{\fluency{}}, and use a simple \textit{length} metric to rate the parsimony of explanations. We also add an additional metric we call \textit{\contextawareness{}}, which considers the unique ability of LLMs to include additional contextual information ``explaining explanations'', beyond that provided by the base explanation.

Our five resulting metrics are described in Table \ref{tab:explanation_metrics}, and are scored on a scale from 0 to 2 (except length). We note that metrics such as \fluency{} are subjective and should be scored by multiple judges. For the work-in-progress experiments described in this paper, the authors rated the responses themselves.

\subsection{Prompt Design Initial Results}
We conducted experiments on two datasets: the student performance dataset \cite{cortez_using_2008} and the Ames housing dataset \cite{de_cock_ames_2011}. We generated SHAP explanations on three instances (input rows) from each dataset, and selected the top three most contributing features as input to the LLM. We then combined these inputs with our five prompts, asking the LLM to generate readable narratives with different parameters\footnote{Details on our experimentation and results can be found at: https://github.com/sibyl-dev/Explingo}. For each of our two test models (GPT-3.5 and GPT-4), we generated and evaluated 90 explanation narratives (5 prompts $\times{}$ 2 datasets $\times{}$ 3 instances $\times{}$ 3 distinct responses). Table~\ref{tab:metric-results} shows the average scores obtained for each model across all the prompts.


GPT-3.5 gives shorter and more fluent responses than GPT-4. GPT-4 performs better on \accuracy{}, \completeness{} and \contextawareness{}. Notably, GPT-4.0's explanations included few errors; however, in real-world high-stakes decision-making contexts, we believe \accuracy{} would need to be consistently scored at 2.


\begin{table}[t]
\caption{Average ratings on our metrics across all prompts.} \label{tab:metric-results}
\centering
\resizebox{\linewidth}{!}{\begin{tabular}{lccccc}
\toprule
\textbf{Model} & \textbf{\Accuracy{}} & \textbf{\Fluency{}} & \textbf{\Completeness{}} & \begin{tabular}{@{}c@{}}\textbf{Context-} \\ \textbf{Awareness}\end{tabular} & \textbf{Length} \\
\midrule
GPT-3.5 & 1.211 & 0.811 & 1.422 & 0.522 & 380.611 \\
GPT-4  & 1.789 & 0.778 & 1.700 & 0.889 & 793.122 \\
\bottomrule
\end{tabular}}
\end{table}

\subsection{User Study}
To understand how people perceive narrative-based explanations versus traditional explanations, we ran a pilot user study. We recruited 20 participants from a random global pool through Prolific\footnote{https://www.prolific.com/}.

We selected the best-rated narrative explanations from our prompt design experiment and generated corresponding bar graph visualizations, resulting in two explanation types (narrative and plot-based). %
Each participant was shown 12 explanations (2 datasets $\times{}$ 3 instances $\times{}$ 2 types). For each explanation, we asked six subjective questions about the user's reception of the explanation, answered on a five-point Likert-type scale. The questions selected were inspired by existing literature for rating explanations and systems \cite{brooke1996sus, hoffman_measures_2023}, specifically chosen to relate to the perceived usability or clarity of explanations. Table \ref{tab:user-study-summary} summarizes the responses to these questions. 

At the end of the study, we asked participants to compare the two explanation types overall. Twelve participants reported preferring narrative-based explanations overall, compared to four for plot-based explanations (the remaining reported no preference). Eleven participants found narrative-based explanations easier to understand compared to five for plot-based, and twelve found narrative-based explanations more informative, compared to six for plot-based.

Overall, our findings motivate the potential benefits of an effective method for generating narrative explanations.

\begin{table}[t]
\centering
\caption{Summary of results from user study. Each rating is converted as strongly disagree=1, strongly agree=5, and then averaged. Narrative-based explanations were preferred on average on all questions.}
\label{tab:user-study-summary}
\resizebox{\linewidth}{!}{\begin{tabular}{@{}lll@{}}
\toprule
                                             & \multicolumn{2}{l}{\textbf{Average Rating}} \\
\textbf{Question}                            & \textbf{Narrative}      & \textbf{Plot}     \\ \midrule
From the explanation, I understand how the algorithm works.                    & \textbf{4.44} & 4.07 \\
The explanation is useful                    & \textbf{4.54}           & 4.07              \\
The explanation helps me know when I should trust the algorithm. & \textbf{3.86} & 3.09 \\
The explanation is clear and understandable. & \textbf{4.55}           & 3.57              \\
The explanation is sufficiently detailed.    & \textbf{4.17}           & 2.99              \\ \bottomrule
\end{tabular}}
\end{table}
\section{Conclusions}
\label{sec:conclusions}
In this paper, we have laid the groundwork for exploring the use of LLMs in XAI by focusing on transforming ML explanations into natural, human-readable narratives. 
By establishing metrics for evaluating narrative explanations and experimenting with prompt-design, we demonstrated the ability of LLMs, particularly GPT-4, to generate sound, complete, and context-aware explanation narratives.

Our pilot user study found that a majority of participants favored narrative-based explanations, finding them easier to understand and more informative. This provides promising insights into the potential benefits of adopting LLM-based narrative explanations.

Moving forward, we aim to further investigate the proposed research directions, such as 
using fine-tuning methods, trying additional LLMs, and exploring the integration of training data and external guides to create context-aware explanations. Future work will also include a larger user study for a more comprehensive evaluation of narrative-based explanations compared to traditional methods.

We offer the following concrete takeaways for the advancement of XAI.
First, the adoption of LLM-based narrative explanations can lead to enhanced user understanding of ML outputs. Second, the superior performance of GPT-4 underscores the advancements in LLM capabilities and their potential impact on XAI. Finally, there remain many promising directions for future research to enhance the quality of narrative explanations.



Ultimately, our research contributes to the ongoing trend of making AI systems more transparent, interpretable, and usable. By providing users with narrative-based explanations, we aim to foster trust and understanding in AI technologies, thereby promoting their responsible and ethical use in diverse domains.







\balance{} 

\bibliographystyle{SIGCHI-Reference-Format}
\bibliography{biblio}

\end{document}